\documentclass[conference,10pt]{IEEEtran}

\usepackage{stfloats}
\usepackage{multirow}
\usepackage[cmex10]{amsmath}
\usepackage{amssymb}
\usepackage{algorithm}
\usepackage{array}
\usepackage{mdwmath}
\usepackage{mdwtab}
\usepackage{fixltx2e}
\usepackage{url}
\usepackage{cmap}
\usepackage{animate}
\usepackage{xspace}
\usepackage{xcolor}
\usepackage{graphicx}
\usepackage[caption=false,font=footnotesize]{subfig}
\usepackage{algpseudocode}
\usepackage{citesort}

\newcommand{\K}{\mathcal{K}}
\newcommand{\gpK}{\kappa_\theta}
\newcommand{\q}{q}

\newcommand{\x}{x}

\newcommand{\ii}{{(i)}}

\begin{document}
\title{Marginalizing Gaussian Process Hyperparameters\\Using Sequential Monte Carlo}

\author{\IEEEauthorblockN{Andreas Svensson$^{\star}$, Johan Dahlin$^{\star\star}$, Thomas B. Sch\"{o}n$^{\star}$}
\IEEEauthorblockA{{\it $^\star$Department of Information Technology, Uppsala University, Sweden}\\
{ \{andreas.svensson, thomas.schon\}@it.uu.se}\\
{\it $^{\star\star}$Division of Automatic Control, Link\"{o}ping University, Sweden}\\
{ johan.dahlin@liu.se}
}}

\maketitle

\begin{abstract}
Gaussian process regression is a popular method for non-parametric probabilistic modeling of functions. The Gaussian process prior is characterized by so-called hyperparameters, which often have a large influence on the posterior model and can be difficult to tune. This work provides a method for numerical marginalization of the hyperparameters, relying on the rigorous framework of sequential Monte Carlo. Our method is well suited for online problems, and we demonstrate its ability to handle real-world problems with several dimensions and compare it to other marginalization methods. We also conclude that our proposed method is a competitive alternative to the commonly used point estimates maximizing the likelihood, both in terms of computational load and its ability to handle multimodal posteriors.
\end{abstract}
\IEEEpeerreviewmaketitle

\section{Introduction}
The Gaussian process (GP) is a non-parametric probabilistic model that can be used to model an unknown nonlinear function $f(\cdot)$ from observed input data $\x$ and (noisy) output data $y = f(\x)$. No explicit form of $f(\cdot)$ is assumed, but some assumptions on $f(\cdot)$ are encoded through the GP prior and a mean function $m_\theta(x)$, a covariance function $\gpK(x,x')$, and their so-called hyperparameters~$\theta \in \Theta$. In mathematical terms, $f$ is a priori modeled to be distributed as
\begin{equation}
f(x) \sim \mathcal{GP}\Big(m_\theta(x),\gpK(x,x')\Big),\label{eq:gp}
\end{equation}
i.e., an infinite-dimensional Gaussian distribution. See \cite{RW:2006} for a more general introduction to GPs.

The posterior distribution over $f(\cdot)$ given data $(y,x)$ is also a GP. This is due to the conjugacy property of the Gaussian distribution. 
The posterior is often greatly influenced by the choice of hyperparameters~$\theta$, which typically are unknown. We therefore propose a method to \emph{marginalize} the hyperparameters in GPs. Marginalization can be seen as averaging over the range of hyperparameters supported by the data and by the prior; $\theta$ can be integrated out by treating it as a random variable with prior $p(\theta)$ and likelihood $p(y|\x,\theta)$, giving rise to the posterior $p(\theta|y,\x)  \propto p(y|\x,\theta)p(\theta)$. For example, the predictive distribution is computed by
\begin{equation}
p(y^*|x^*,y,\x) = \int\! p(y^* |x^*,y,\x,\theta)p(\theta|y,\x)\textrm{d}\theta,\label{eq:bpred}
\end{equation}
which unfortunately is analytically intractable. However, using a Monte Carlo method to obtain $N$ (weighted) samples $\{w^\ii, \theta^\ii\}_{i=1}^N$ of the distribution $p(\theta|y,\x)$, the predictive distribution \eqref{eq:bpred} can be approximated by
\begin{equation}
\widehat{p}(y^*|x^*,y,\x) = \sum_{i=1}^N w^\ii p(y^* |x^*,y,\x,\theta^\ii),\label{eq:bpreda}
\end{equation}
where the weights are normalized, i.e., $\sum_i w^\ii = 1$.

\begin{figure}[t!]
\centering
\captionsetup[subfigure]{width=\columnwidth}
\subfloat[\label{fig:intro_eb}Gaussian process regression for the data set defined by the red dots, using two different point estimates for the hyperparameters, each corresponding to a local minimum in (b, left).]{\includegraphics[height=80pt]{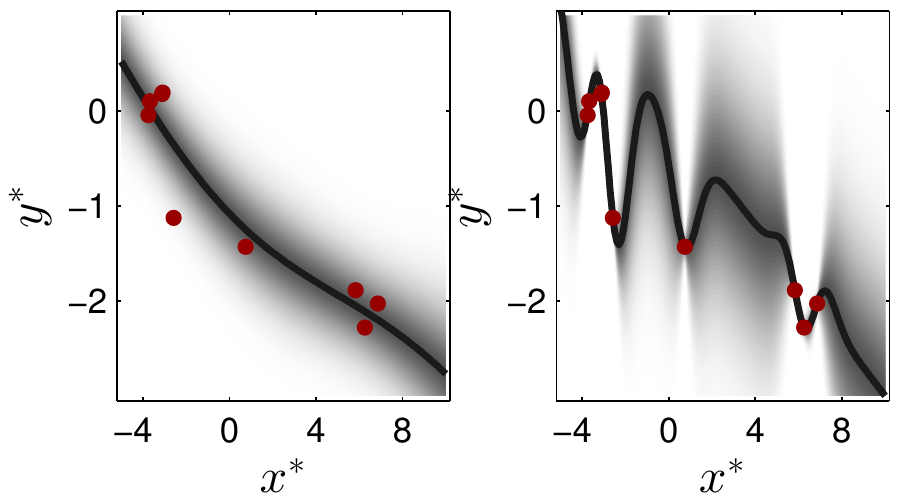}\hspace{2.8cm}}\\
\subfloat[\label{fig:intro_smc}Left: the (multimodial) hyperparameter posterior conditional on the 9 data points. Right: the posterior using the proposed method (which marginalizes the hyperparameters, and thus handles the multimodality).]{\includegraphics[height=80pt]{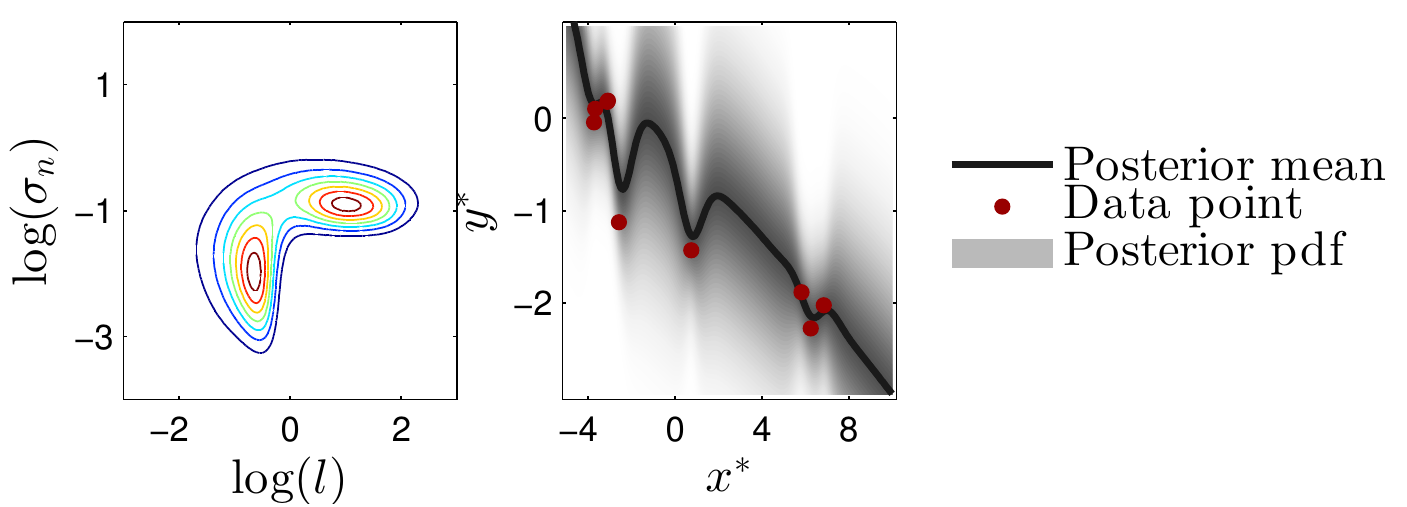}}
\caption{A small example illustrating the influence of the hyperparameters in the GP prior to the posterior estimate.}
\label{fig:intro}
\vspace{-0.2cm}
\end{figure}

A common alternative to marginalization is to choose a point estimate of $\theta$ using an optimization procedure maximizing the likelihood $p(y|\x,\theta)$ (sometimes referred to as empirical Bayes). This may be difficult if the likelihood is multimodal. See the small toy example in Figure~\ref{fig:intro} illustrating the robustness of marginalization compared to point estimates. There are also situations where point estimates are not sufficient, and marginalization is necessary, such as the change point detection problem in Section~\ref{sec:application}.

Our contribution is a method for sampling from the hyperparameter posterior distribution $p(\theta|y,\x)$, based on sequential Monte Carlo (SMC) samplers \cite{DDJ:2006}. SMC samplers and their convergence properties are well studied \cite{Whiteley:2012}. 

Several methods have previously been proposed in the literature for marginalization of the GP hyperparameters: Bayesian Monte Carlo (BMC) \cite{ORR+:2008}, slice sampling \cite{AG:2005}, Hamiltonian Monte Carlo \cite{Neal:2010,STR:2010}, and adaptive importance sampling (AIS) \cite{PGS:2014}. Particle learning which is closely related to SMC has been proposed by \cite{GP:2011} for this purpose. The work by \cite{GP:2011}, however, is not targeting the hyperparameters directly, and makes (possibly restrictive) assumptions on conjugate priors and model structure.

In this paper, we compare our proposed method to some of these methods, and apply it to two real-data problems: the first demonstrates that marginalization does not have to be more computationally demanding than finding point estimates. The second example, which deals with a fault detection problem from industry, is possible only with an efficient method for marginalization. Our proposed method (and all examples) are available as Matlab code via the first authors homepage. 

From the experiments, we conclude that the advantages of the proposed method are (i) robustness towards multimodal hyperparameter posteriors, (ii) simplified tuning (compared to some other alternatives), (iii) competitive computational load, and (iv) online updating of hyperparameters as the data record grows.

\section{Sampling hyperparameters using SMC}\label{sec:SMC}
For the numerical marginalization~\eqref{eq:bpreda}, we require $N$ samples, known as \emph{particles}, from the posterior. In this section, we discuss how to use a SMC sampler \cite{DDJ:2006} to generate such a particle system $\{\theta^\ii, w^\ii\}_{i=1}^N$, where $w^\ii$ is the weight of particle $\theta^\ii$. The underlying idea is to construct a sequence of probability distributions ($\{\pi_0, \dots, \pi_P\}$), starting from the prior, and ending up in the posterior. The particles are then `guided' through the sequence.

To construct a sequence $\{\pi_0, \dots, \pi_P\}$, we use the fact that $p(\theta|y,\x)$ depends on the data $(y,\x)$, by partitioning the data points into $P$ disjoint batches $\{B_n\}_{n=1}^P$ and adding them sequentially as $\pi_n(\theta) \propto p(y_{B_{1:n}}|\x_{B_{1:n}},\theta)p(\theta)$.

To guide the particles through the smooth sequence $\{\pi_0, \dots, \pi_P\}$, we will iteratively apply the three steps weighting, resampling and propagation, akin to a particle filter.

In the \emph{weighting} step, the `usefulness' of each particle is evaluated. To ensure convergence properties, the particles can be evaluated as \cite[Section 3.3.2]{DDJ:2006}
\begin{equation}
w^\ii_n = \frac{\pi_n(\theta^\ii_{n-1})}{\pi_{n-1}(\theta^\ii_{n-1})}w^\ii_{n-1}. \label{eq:w}
\end{equation}

To avoid numerical problems, the particles have to be \emph{resampled}. The idea is to duplicate particle with large weights, and discard particles with small weights. 

To \emph{propagate} the particles $\theta_{n-1}^\ii$ from $\pi_{n-1}$ to $\pi_{n}$, a Metropolis-Hastings (MH) kernel $\K: \Theta \mapsto \Theta$ with invariant distribution $\pi_n$ can be used. The procedure of propagating $\theta_{n-1}$ (a sample of $\pi_{n-1}$) to $\theta_{n}$ (a sample of $\pi_{n}$) by $\K$ is as follows:
(i)~Sample a new particle  $\theta'$ from a proposal $\q(\cdot|\theta_{n-1})$, e.g., a random walk with variance~$h$. (ii)~Compensate for the discrepancy between $\pi_n$ and $\q$ by setting $\theta_{n} = \theta'$ with probability
\begin{equation}
\alpha(\theta_{n},\theta') = \min \left\{1, \tfrac{\pi_n(\theta')}{\pi_n(\theta_{n})} \tfrac{\q(\theta_{n}|\theta')}{\q(\theta'|\theta_{n})}\right\}, \label{eq:alpha}
\end{equation}
and otherwise $\theta_{n} = \theta_{n-1}$. To improve the mixing, this procedure can be repeated $K$ times. For this, we use the notation $\theta_{n-1} = \theta_{n}^0 \rightarrow \theta_{n}^1 \rightarrow \dots \rightarrow \theta_{n}^K = \theta_n$.

\begin{figure}[t]
	\vspace{2mm}
\subfloat[\label{fig:intro1}A transition from the prior $p(\theta)$ to the posterior $p(\theta|y,\x)$ for the data in Figure~\ref{fig:intro_smc}, obtained by adding 3 data points in each step to the likelihood. The particles are obtained from the SMC sampler.]{
\includegraphics[width=0.22\columnwidth]{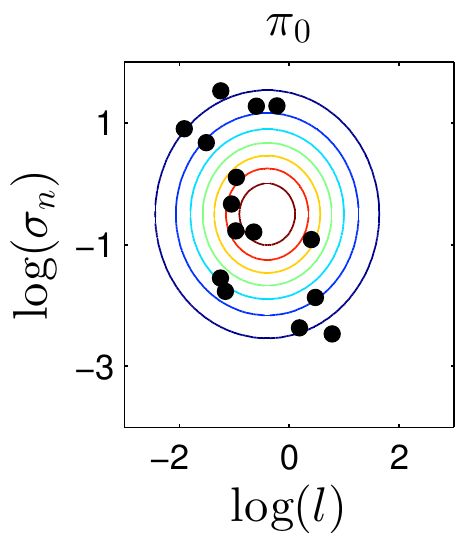}
\includegraphics[width=0.22\columnwidth]{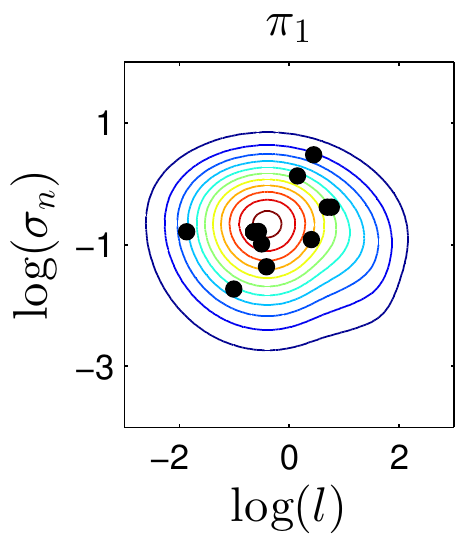}
\includegraphics[width=0.22\columnwidth]{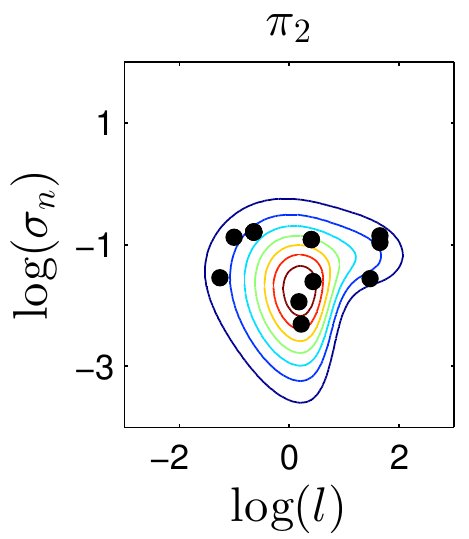}
\includegraphics[width=0.22\columnwidth]{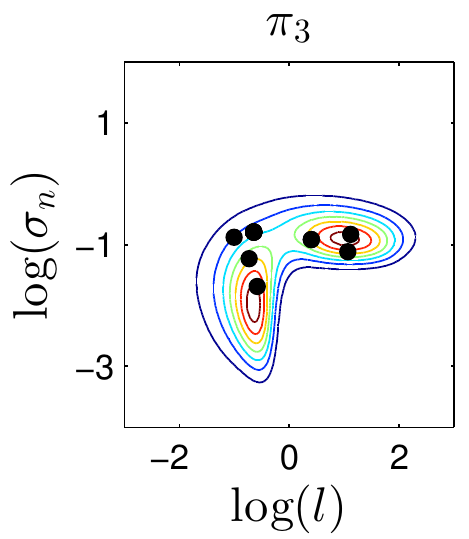}
}\\
\subfloat[\label{fig:intro2}GP regression with marginalized hyperparameters from the corresponding posterior, obtained as a by-product from the particles depicted in (a). From left to right, 0 data points (i.e., the prior), 3 data points, 6 data points, and 9 data points. As we formulated the problem, only the rightmost figure is of interest. This illustrates however how this method can be used in online problem in a natural way.]{
\includegraphics[width=0.22\columnwidth]{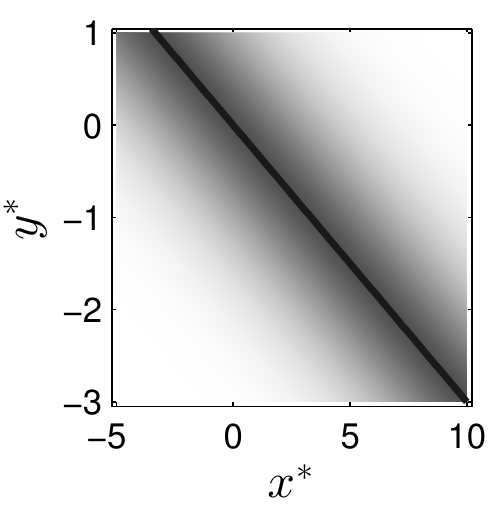}
\includegraphics[width=0.22\columnwidth]{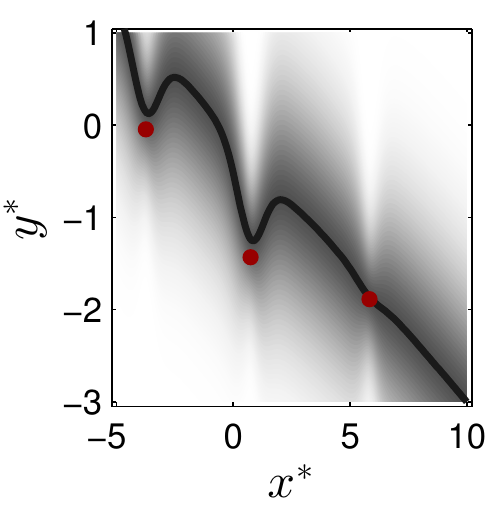}
\includegraphics[width=0.22\columnwidth]{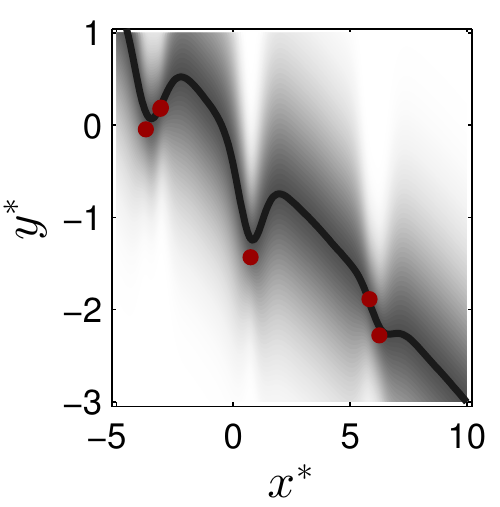}
\includegraphics[width=0.22\columnwidth]{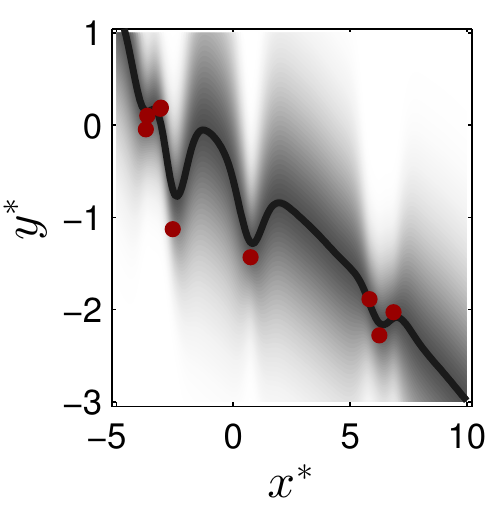}
}
\caption{Illustration of the SMC sampler, as it evolves from the prior (no data) to the posterior (all data).}
\label{fig:intro3}
\vspace*{-2em}
\end{figure}

\begin{algorithm}[t]
	\caption{Hyperparameter posterior sampler}
	\small
	\begin{algorithmic}[1]
		\Require Data $(y,\x)$, GP prior, and prior $p(\theta)$.
		\Ensure $N$ samples $\{\theta^\ii\}_{i=1}^N$ from $p(\theta|y,x) \propto p(y|\x,\theta)p(\theta)$.
		\vspace{2pt}
		\Statex \emph{All statements with superscript $\ii$ are for $i = 1, \dots, N$.}
		\vspace{2pt}
		\State Define $\pi_n(\theta) = p(y_{B_{1:n}}|\x_{B_{1:n}},\theta)p(\theta)$ by partitioning the data into $P$ batches $\{B_n\}_{n=1}^P$.\label{alg1:pi}
		\State Sample $\theta^\ii_0$ from $p(\theta)$ ($=\pi_0(\theta)$).\label{alg1:init}
		\For{$n = 1$ to $P$}\label{alg1:Pfor}
		\State Update weights according to \eqref{eq:w}.\label{alg1:weights}
		\State Resample $\{\theta^\ii_n,w^\ii_n\}_{i=1}^N$ if needed.
		\For{$k = 1$ to $K$}\label{alg1:Kfor}
		\State Propose $\theta'^\ii$ from $q(\theta'|\theta_n^{k-1,\ii})$.
		\State Set $\theta^{k,\ii}_{n} = \theta'^\ii$ with prob. $\alpha(\theta_{n}^{k-1,\ii},\theta'^\ii)$ \eqref{eq:alpha}.
		\EndFor\label{alg1:endKfor}
		\EndFor\label{alg1:endPfor}
	\end{algorithmic}
	\label{alg:smcs-gp}
\end{algorithm}

We now have an SMC sampler to obtain samples from the hyperparameter posterior, summarized in Algorithm~\ref{alg:smcs-gp} and illustrated by Figure~\ref{fig:intro3}. From the figure, the suitability to online applications is clear: If another data point is added to the data, the sequence can be extended to $\pi_4$ including the new data point, and only the transition from $\pi_3$ to $\pi_4$ has to be performed.

We make use of the adaptive SMC sampler by \cite{FT:2013} in the numerical examples to adapt the proposal $q$ automatically.

The computational cost of Algorithm~\ref{alg:smcs-gp} is in practice governed by the $2NPK$ evaluations of the likelihood $p(y|\x,\theta)$. Hence, it is important to choose the number of samples $N$, SMC steps $P$, and MH-moves per SMC-step $K$ sensibly. An idea of sensible numbers will be given along with the examples in the next section.

\pagebreak

\section{Examples and results}\label{sec:numex}

We consider three examples for demonstrating our proposed approach. First, we consider a small simulated example, also comparing to alternative sampling methods, and thereafter two applications with real-world data. The first real-data example is a benchmark problem to compare the marginalization approach in Algorithm~\ref{alg:smcs-gp} to the point estimates obtained using optimization. In the third example, we illustrate how we can make use of our solution within a GP-based online change point detection algorithm. To this end, we require marginalization of the hyperparameters, so an efficient hyperparameter posterior sampler is indeed a key enabler for this. The online nature of the problem also fits well to the possibility to update the samples in Algorithm~\ref{alg:smcs-gp} online, as discussed in Section~\ref{sec:SMC}.

\subsection{Simulated example}
We consider a small problem of 5 data points, and a covariance and mean function with 7 hyperparameters in total. We begin by considering the problem of marginalizing out 7 hyperparameters in a GP prior given 5 data points. Here, we are interested in comparing the performance of our SMC sampler (Algorithm~\ref{alg:smcs-gp}) with some popular alternative methods; BMC \cite{ORR+:2008}, AIS \cite{PGS:2014}, and (deterministic) griding.

\begin{figure}[t]
	\centering
	\includegraphics[width=\columnwidth]{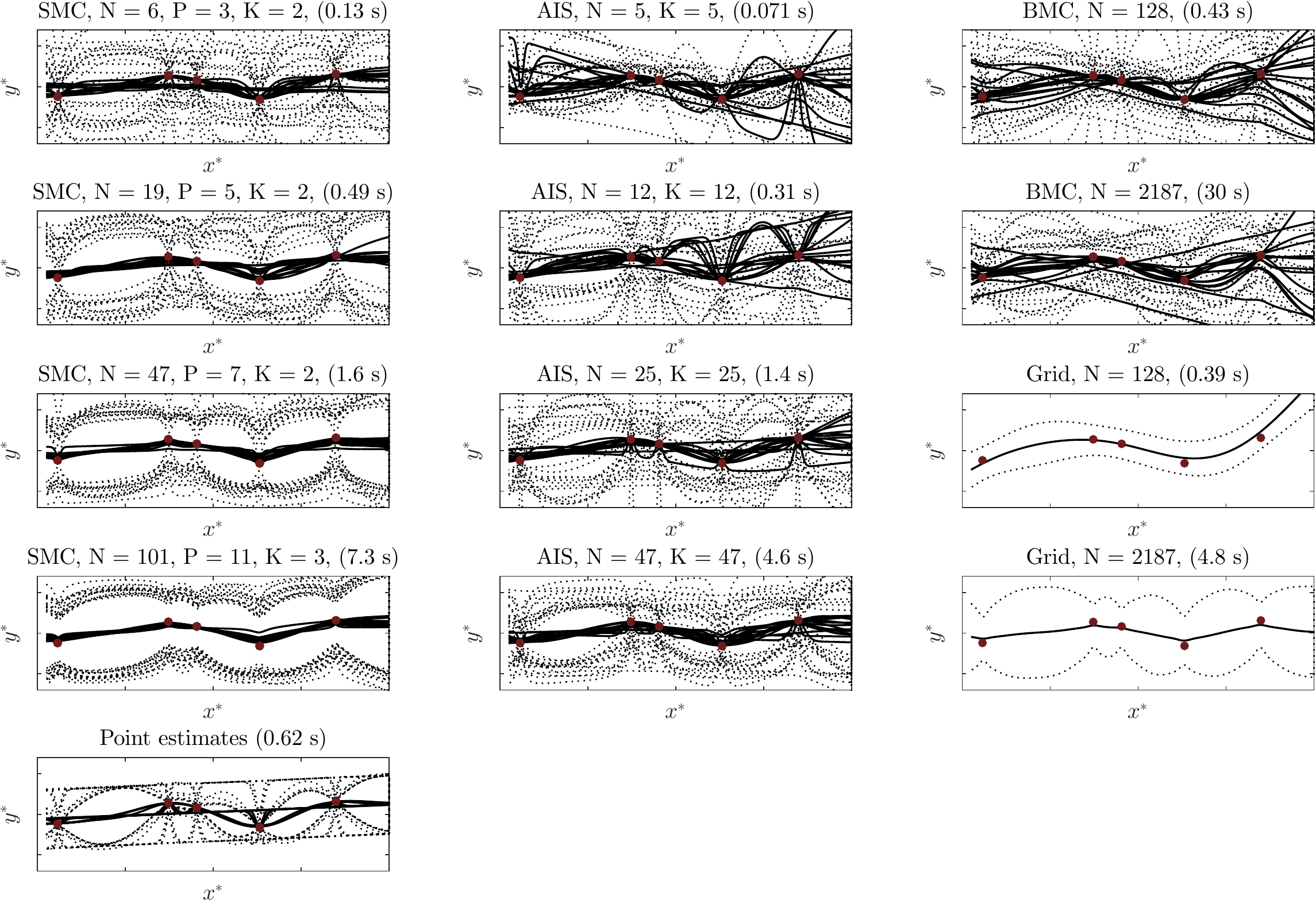}
	\vspace*{-5ex}
	\caption{Comparison between 15 runs of SMC (Algorithm~\ref{alg:smcs-gp}), BMC, AIS, and griding, as well optimized point estimates. The predictions (mean, solid, and 3 standard deviations, dashed) are shown, together with the red data points. The number of particles/samples/grid points is denoted by $N$, while $K$ and $P$ are algorithm specific tuning parameters. The mean computation time is also shown. All axis are equally scaled. \newline \newline The quite `messy' look in most of the plots indicates that the same method (with fixed settings) behaves differently on each run, which of course is an unwanted effect. However, the SMC sampler is not suffering from this problem for $N,P,K$ large enough. This effect should also be expected for AIS and BMC, but apparently they need more samples/iterations (and thus computing time) than presented here before that effect can be seen.}
	\label{fig:comparison}
	\vspace{-0.8em}
\end{figure}

The results for 15 runs are presented in Figure~\ref{fig:comparison}; it is indeed good if the variance between consecutive runs of the same algorithm gives similar results. The variations between the runs decrease faster for Algorithm~\ref{alg:smcs-gp} than for the comparable methods. When the GP prior has few hyperparameters, we conclude that the AIS and griding might be competitive methods. We have not managed to obtain competitive results with BMC for any problem size, but it should be noted that the computational load of BMC can be substantially decreased if the hyperparameter prior is independent between the dimensions.

The results for the conceptually different point estimates are also presented in Figure~\ref{fig:comparison}. The initialization point to the optimization algorithm is drawn from the prior: although it is a deterministic method, it is obviously very sensitive to the initialization.

\subsection{Learning a robot arm model}\label{sec:sarcos}

We consider the problem of learning the inverse dynamics of a seven degrees-of-freedom SARCOS antromorphic robot arm \cite{VS:2000,RW:2006}. We use the same setup as \cite[Section 2.5]{RW:2006}, i.e., a non-trivial setting involving 23 hyperparameters.

To handle the size of the data set ($44\,484$ training and $4\,449$ test data points), we make use of a subset of: (i) datapoints and (ii) regressors as discussed by  \cite[Section 8.3.1]{RW:2006}. To use our method, we sample the hyperparameters from the posterior with a subset of $m$ data points. For comparability, we have also reproduced the results using point estimates from \cite{RW:2006}. The results are reported in Table~\ref{table:sarcos}. For Algorithm~\ref{alg:smcs-gp}, $N = 15$, $P = 20$ and $K = 5$ was used. The priors to the logarithms of the length-scale and the signal variance are $\mathcal{N}(3,3)$, and $\mathcal{N}(1,1)$ for the noise variance.

\pagebreak
\begin{table}[t]
\caption{Results for the SARCOS example in Section~\ref{sec:sarcos}.}
\label{table:sarcos}
\scriptsize
\begin{sc}
\begin{tabular}{lllll}
Method & $m$ & SMSE $(\times 10^{-2})$ & MSLL & Time (s) \\
\hline
\multicolumn{3}{l}{Subset of datapoints}  \\
Point est.& 256 & 8.36 $\pm$ 0.80 & -1.38 $\pm$ 0.04 & 6.8 \\
\bf{SMC}  & \bf{256} & \bf{8.10 $\pm$ 1.32} & \bf{-1.38 $\pm$ 0.56} & \bf{7.1} \\
Point est.& 512 & 6.36 $\pm$ 1.13 & -1.51 $\pm$ 0.05 & 26.4 \\
\bf{SMC}  & \bf{512} & \bf{6.13 $\pm$ 0.91} & \bf{-1.49 $\pm$ 0.04} & \bf{22.3}\\
Point est.& 1024& 4.31 $\pm$ 0.16 & -1.66 $\pm$ 0.02 & 101\\ 
\bf{SMC}  & \bf{1024}& \bf{4.54 $\pm$ 0.33} & \bf{-1.61 $\pm$ 0.03} & \bf{92.5}\\
Point est.& 2048& 2.99 $\pm$ 0.08 & -1.78 $\pm$ 0.03 & 423\\ 
\bf{SMC}  & \bf{2048}& \bf{3.33 $\pm$ 0.28} & \bf{-1.69 $\pm$ 0.06} & \bf{405}\\
		  \hline 
\multicolumn{3}{l}{Subset of regressors} \\
Point est.& 256 & 3.67 $\pm$ 0.17 & -1.63 $\pm$ 0.02  & 6.8\\
\bf{SMC}  & \bf{256} & \bf{3.55 $\pm$ 0.28} & \bf{-1.65 $\pm$ 0.05} & \bf{7.1} \\
Point est.& 512 & 2.77 $\pm$ 0.44 & -1.79 $\pm$ 0.07 & 26.4\\
\bf{SMC}  & \bf{512} & \bf{2.89 $\pm$ 0.20} & \bf{-1.77 $\pm$ 0.03} & \bf{22.3}\\
Point est.& 1024& 2.03 $\pm$ 0.11 & -1.95 $\pm$ 0.03 & 101\\ 
\bf{SMC}  & \bf{1024}& \bf{2.00$^\star$}    & \bf{-1.95$^\star$}      & \bf{92.5}\\ 
\hline
\end{tabular}
\end{sc}
\vspace*{-1em}
\end{table}

Table~\ref{table:sarcos} presents the results in the same way as \cite[Table 8.1]{RW:2006}. SMSE is the standardized mean square error (i.e., mean square error normalized by the variance of the target), and MSLL is the mean standardized log loss; 0 if predicting using a Gaussian density with mean and variance of the training data, and negative if `better'. The time is referring to the time required to sample and optimize the hyperparameters, respectively (not including the test evaluation). Numerical problems were experienced for large $m$, therefore $^\star$ indicates runs where no interval can be reported.

Table~\ref{table:sarcos} indicates no significant difference between the performance of our method and point estimates. It is however worth also to note the computational load: As Algorithm~\ref{alg:smcs-gp} apparently makes an equally good job in finding relevant hyperparameters as the optimization, it is a confirmation that our proposed method is indeed a competitive alternative to point estimates even for large problems.

\pagebreak

\subsection{Fault detection of oxygen sensors}\label{sec:application}

We now consider data from the wastewater treatment plant K\"{a}ppalaverket, Sweden.
An oxygen sensor measures the dissolved oxygen (in mg/l) in a bioreactor, but the sensor gets clogged because of suspended cleaning. The identification of such events is relevant to the control of wastewater treatment plants \cite{OCC+:2014}. We apply the GP-based online change point detection algorithm by \cite{STR:2010}, where the hyperparameters are marginalized using our proposed method.

\begin{figure}
\centering
\subfloat[\label{fig:cpd_res}Measurements of dissolved oxygen (in mg/l) in a bioreactor with a sampling period of 15 minutes. The indicated change points are marked in red. Especially as the algorithm is fully Bayesian, the outcome is one probability distribution per data sample. This is comprehensively illustrated as the occurrence of change points in `backwards simulations' through these distributions. A more intensive red color is a more likely change point.]{\includegraphics[width=\columnwidth]{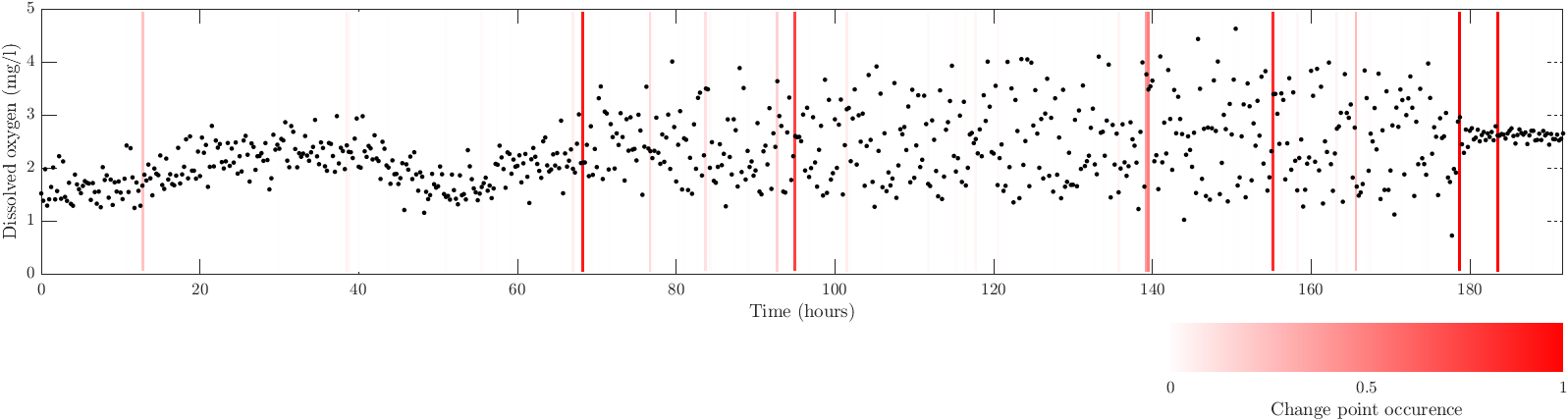}}\\
\subfloat[\label{fig:cpd_segm}Thresholding of Figure (a), with GP regression in each obtained segment. The different characteristics in different segments are possible due to marginalization of the hyperparameters.]{\includegraphics[width=\columnwidth]{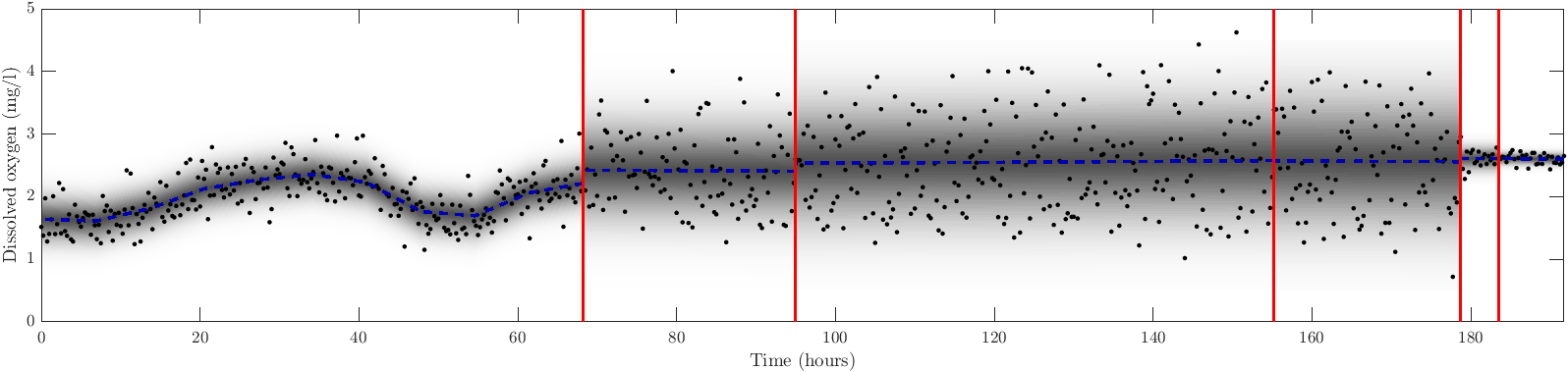}}
\caption{Results for the GP-based change point detection.}
\end{figure}

The GP-based change point detection presented by \cite{STR:2010} can be summarized as follows: If data $y_{1:T}$ undergo a change at time $r$, it is of interest to (online) detect $r$, i.e., estimate $p(r|y_{1:t})$. The algorithmic idea is a recursive message passing scheme, updating the probability $p(r_t, y_{1:t})$, where $r_t \in \{1, \dots, t\}$ is the last change point at time $t$.

To make predictions using a GP model, the hyperparameters either have to be fixed across all data segments, or marginalized. As it is not relevant to use fixed hyperparameters, an efficient sampling algorithm is a key enabler in solving this problem. The consecutive predictions $p(y_t|r_{t-1},y_{r_{t}:t-1})$ and $p(y_{t+1}|r_{t-1},y_{r_{t}:t})$ are both needed for the algorithm, hence our approach fit this problem well, as discussed in Section~\ref{sec:SMC}. We used $N = 25$ particles. On average, sampling the hyperparameters, i.e., one run of Algorithm~\ref{alg:smcs-gp}, took 0.55 seconds on a standard desktop computer.

The results are presented in Figure~\ref{fig:cpd_res}. The expected points, suspension and resuming of the cleaning, are indeed indicated. An interpretation of the result is obtained by converting the results to point estimates by thresholding, and plotting at the GP regression for each individual segment, see Figure~\ref{fig:cpd_segm}. 

Note the data-driven nature of the algorithm, as no explicit model of the sensor was used at all. The tuning parameters are the covariance and mean functions, the prior of the change points and the hyperparameter priors.

\section{Conclusion}

We have proposed and demonstrated an SMC-based method to marginalize hyperparameters in GP models. The observed benefits are robustness towards  multimodal posteriors (Figure~\ref{fig:intro}) and a competitive computational load (Section~\ref{sec:sarcos}), also compared to the commonly used point estimates of the hyperparameters. We have been able to cope with a hyperparameter space of dimension 23 (Section~\ref{sec:sarcos}), and also concluded a sound convergence behavior (Section~\ref{sec:numex}). Finally, the online update of the hyperparameters has been shown useful within the industry-relevant data-driven fault detection application (Section~\ref{sec:application}). As a future direction, it would be interesting to apply our method to the challenging GP optimization problem of system identification \cite{DL:2014}.

\section*{Acknowledgments} 
 
\footnotesize
This work was supported by the project \emph{Probabilistic modeling of dynamical systems} (Contract number: 621-2013-5524) funded by the Swedish Research Council (VR). We would also like to thank Oscar Samuelsson and Dr. Jes\'{u}s Zambrano for providing the sensor data in Section~\ref{sec:application}.

\bibliographystyle{IEEEtran}
\bibliography{IEEEabrv,../../../../references}

\end{document}